\documentclass{article}

\usepackage[preprint, nonatbib]{neurips_data_2024}
\usepackage[utf8]{inputenc} % allow utf-8 input
\usepackage[T1]{fontenc}    % use 8-bit T1 fonts
\usepackage{hyperref}       % hyperlinks
\usepackage{url}            % simple URL typesetting
\usepackage{booktabs}       % professional-quality tables
\usepackage{amsfonts}       % blackboard math symbols
\usepackage{nicefrac}       % compact symbols for 1/2, etc.
\usepackage{microtype}      % microtypography
\usepackage{xcolor}         % colors
\usepackage{comment}
\usepackage{multirow}
\usepackage{graphicx}
\usepackage{pifont}
\usepackage{amsmath}
\usepackage{booktabs}
\usepackage{subcaption}
\hypersetup{
    colorlinks=true,
    urlcolor=cyan,
}

\title{SegFaults in the Earth: Leveraging Crowdsourcing to
analyze Subsurface Images}
\title{DeFaults Dataset: Delineating Subsurface Faults using Crowdsourcing}
\title{DeFaults Dataset: Leveraging Crowdsourcing to Analyze and Delineate Subsurface Faults}
\title{CRACKS: Crowdsourcing Resources for Analysis and Categorization of Key Subsurface faults}

\author{%
Mohit Prabhushankar$^{1}$, Kiran Kokilepersaud$^{1*}$, Jorge Quesada$^{1*}$, Yavuz Yarici$^1$\thanks{Equal contribution}  , Chen\\ \textbf{Zhou$^1$, Mohammad Alotaibi$^1$, Ghassan AlRegib$^1$, Ahmad Mustafa$^2$, Yusufjon Kumakov$^3$}\\
  $^1$OLIVES at the Centre for Signal and Info. Processing, Georgia Tech, Atlanta, GA 30332, USA\\
  $^2$Occidental Petroleum, Houston, Texas, USA\\
  $^3$Institute of Geology and Exploration of Oil and Gas Fields, Uzbekistan\\
  \texttt{\{mohit.p, kpk6, yavuzyarici, jpacora3, chen.zhou, malotaibi44,} \\
  \texttt{alregib\}@gatech.edu}\\
  \texttt{ahmad\_mustafa@oxy.com}\\
  \texttt{kumakovyusufjon@gmail.com}\\
}
\begin{document}

\maketitle

\begin{abstract}

Crowdsourcing annotations has created a paradigm shift in the availability of labeled data for machine learning. Availability of large datasets has accelerated progress in common knowledge applications involving visual and language data. However, specialized applications that require expert labels lag in data availability. One such application is fault segmentation in subsurface imaging. Detecting, tracking, and analyzing faults has broad societal implications in predicting fluid flows, earthquakes, and storing excess atmospheric CO$_2$. However, delineating faults with current practices is a labor-intensive activity that requires precise analysis of subsurface imaging data by geophysicists. In this paper, we propose the \texttt{CRACKS} dataset to detect and segment faults in subsurface images by utilizing crowdsourced resources. We leverage Amazon Mechanical Turk to obtain fault delineations from sections of the Netherlands North Sea subsurface images from (i) $26$ novices who have no exposure to subsurface data and were shown a video describing and labeling faults, (ii) $8$ practitioners who have previously interacted and worked on subsurface data, (iii) one geophysicist to label $7636$ faults in the region. Note that all novices, practitioners, and the expert segment faults on the same subsurface volume with disagreements between and among the novices and practitioners. Additionally, each fault annotation is equipped with the confidence level of the annotator. The paper provides benchmarks on detecting and segmenting the expert labels, given the novice and practitioner labels. Additional details along with the dataset links and codes are available at \href{https://alregib.ece.gatech.edu/cracks-crowdsourcing-resources-for-analysis-and-categorization-of-key-subsurface-faults/}{our website}.

\end{abstract}

\section{Introduction}
\label{sec:Intro}
\begin{figure}[t]
\centering
\includegraphics[scale =.5]{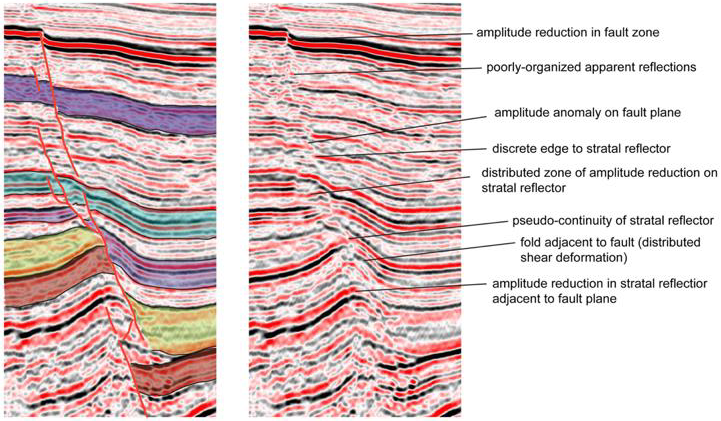}
\vspace{-1.5mm}
\caption{Expert interpretation of a seismic fault.}\label{fig:Concept}\vspace{-3mm}
\end{figure}

Understanding Earth’s subsurface structures via seismic image analysis has been and continues to be an essential component of various applications such as earthquake monitoring, carbon sequestration, and hydrocarbon exploration. Typically, geophysical expert interpreters first assume a geological model based on the analysis of various attributes of seismic data, in addition to the geological history of the region~\cite{alregib2018subsurface}. Thereafter, interpreters segment various subsurface structures of interest. An important subsurface structure that has broad societal impact is faults. Fig.~\ref{fig:Concept} illustrates the process used by an expert interpreter to segment a fault in the earth's subsurface. A fault is defined as a lineament or planar surface across which apparent relative displacement occurs in the rock layers. Visually, the fault is apparent, even to non-geophysicists, as discontinuities in the earth's subsurface. Geophysicists model the dominant structures (colored regions on the left image in Fig.~\ref{fig:Concept}) and check for discontinuities. However, non-geophysicists fall short when segmenting faults in the following cases: (i) it is not always clear where a fault starts and where it ends, and (ii) it is not always clear when there are a series of minor faults or there is one major fault. For instance, consider the cyan segmented region in Fig.~\ref{fig:Concept} as our region of interest (ROI). While the region above and below (in purple) the ROI shows displacement, there is no displacement within the ROI itself. Additionally, the magnitude of displacement within the yellow and red regions below the ROI is much larger than the purple region above the ROI. This indicates two separate faults. Further, the red and yellow regions to the left of the fault are next to each other while they deviate to the right of the fault. This indicates a series of minor faults. Note that this analysis requires geological study and structural segmentation from expert geophysicists.

\textbf{Societal impact} Analyzing and monitoring faults has multifarious societal benefits. Faults may be sources of societal hazard that allow earthquakes, sometimes exacerbated by human activities. Faults in the subsurface can serve as conduits to fluid-flow. Decades of experience has proven that CO$_2$ can be stored and sealed within the earth's subsurface, provided that there are no leakage pathways generated through reactivation of faults. This is termed carbon sequestration and it allows capturing and storing excess atmospheric carbon dioxide with the goal of reducing global climate change~\cite{warwicknational}. However, carbon dioxide may leak back into the atmosphere if there are are unknown non-sealing faults in the injection sites, requiring fault delineation before sequestration.

\textbf{Automating fault delineation} Fault delineation is the combined process of detecting and segmenting faults. As illustrated in Fig.~\ref{fig:Concept}, fault delineation process is time consuming and labor intensive \cite{mustafa2024visual}. Automated and semi-automated interpretation through computational algorithms have been investigated~\cite{wang2016interactive, cohen2006detection, roberts2001curvature, di2017seismic, shafiq2018towards, kokilepersaud2022volumetric}. However, training generalizable machine learning algorithms require labeled fault datasets which is lacking. Seismic surveys have grown over the years in both complexity and data sizes. However, different subsurface regions within the earth have different geologies, all of which cannot be labeled by geophysicists. To tackle this challenging situation, a common practice is to utilize synthetic data to construct deep learning models~\cite{wu2019faultseg3d, li2022automatic, shafiq2018towards, shafiq2018leveraging}. Synthetic data offers the advantage that the exact segmentation of faults is available. However, this is not realistic. Real data suffers from confounding factors that have not yet been modeled synthetically. This is specifically true in faults, where factors like earthquake epicenter depths create significant uncertainties~\cite{vakarchuk2019effect}. 

\textbf{CRACKS Dataset Contributions} To alleviate the lack of expertise, we propose the \texttt{CRACKS} dataset, the first subsurface fault-labeled data utilizing resources from crowdsourcing platforms. Specifically, we provide annotations from multiple expertise levels, of differing confidence, including from novices, practitioners, and an expert. We show that the noisy annotations from novices and practitioners are not trivial and can be utilized to model expert annotations. Annotations from multiple novices on the same image have large disagreements between them and from the expert. This allows \texttt{CRACKS} dataset to be used in a non-traditional way where the same images used during training with novice and practitioner labels can be used during inference to model expert annotations. 
\section{Related Work}
\label{sec:LitReview}
\paragraph{Seismic Data} Seismic images are obtained via transmitting and receiving signals, generally acoustic waves, through the earth's subsurface across a predefined survey area. Seismic image characteristics are dependent on the subsurface geology. Subsurface under the Netherlands North Sea F3 block~\cite{F3_data} has its own unique characteristics, distinct from other regions like Great South Basin in New Zealand~\cite{carter1988post}, and the South China Sea~\cite{chen2011seismic}. In this paper, we annotate faults on the Netherlands F3 block due to the following reasons:

\begin{itemize}
    \item The F3 block, apart from faults, has other dominant structures in the subsurface images. This provides a larger diversity in image features for deep learning algorithms to learn from.
    \item The F3 block is available for use publicly with no additional licensing and has a number of other annotations including structural interpretations in LANDMASS~\cite{alaudah2018structure} and Facies~\cite{alaudah2019machine} annotations. These annotations are not fault related but can be used as pretext tasks~\cite{kokilepersaud2022volumetric}.
    \item The F3 block has been used for a number of machine learning applications including contrastive learning~\cite{kokilepersaud2022volumetric}, active learning~\cite{benkert2023samples, mustafa2021man}, visual explainability~\cite{prabhushankar2020contrastive, alregib2022explanatory}, and model~\cite{benkert2022reliable} and label uncertainty estimation~\cite{zhou2023perceptual} among others.
    \item The faults in F3 block are easier to visually inspect and label as compared to polygonal faults in other parts of the earth's subsurface~\cite{carter1988post, chen2011seismic}. This ensures that specialized seismic software like OpendTect~\cite{dgbpy_2019} is not required for crowdsourced labeling.
\end{itemize}

\paragraph{Leveraging Crowdsourcing for Learning Expert Labels} Crowdsourcing labels from multiple annotators has been a common approach to obtain large labeled data. Crowdsourcing has fostered ML research in common knowledge applications including image classification \cite{peterson2019human, collins2022eliciting} and natural language processing \cite{uma2021learning, poesio2019crowdsourced}. However, specialized applications that require domain expert annotations lag in label availability. Some medical applications involve collecting annotations from multiple observers. However, the number of observers is typically small and the labeling is time-consuming \cite{armato2011lung, litjens2012pattern, almazroa2017agreement, menzequantification}. \texttt{CRACKS} is the first dataset that leverages crowdsourcing resources for delineating expert-labeled faults from annotators with limited and no expertise. For comparison, we summarize various key characteristics of the datasets that leverage crowdsourcing or provide multiple annotations in Table~\ref{tab:stats_datasets}. Note that none of these datasets are specifically for seismic data. Many of these datasets do not provide annotations from real expertise hierarchies. In certain specialized domains such as the medical domain, gold labels are usually derived from the majority voting among a few observers. However, majority voting rarely produces ground truth-expert labels in fault delineations due to various levels of labeling consistency and disagreement across different expertise levels. Even in medical data, arbitration between stakeholders is required~\cite{prabhushankar2022olives}. Compared to other datasets, \texttt{CRACKS} provides real-world settings where crowdsourcing from multiple levels of expertise can be used as an alternative to time-consuming labeling by experts. This unique characteristic of \texttt{CRACKS} facilitates the development of fine-tuning strategies that use limited expert labels to fine-tune ML models trained using a large number of labels from less experienced annotators, such as practitioners and novices. The availability of crowdsourced labels from different expertise levels also encourages benchmarking of self-supervised learning \cite{chen2020simple}. In addition to multiple levels of expertise, \texttt{CRACKS} also provides three categories of annotation confidence, which can be used for weighted or fused labels during training to enhance generalizability. Finally, \texttt{CRACKS} provides multiple tasks of interest including detection and instance segmentation \cite{glenn_jocher_2022_7347926}. To the best of our knowledge, \texttt{CRACKS} is the first subsurface fault dataset that provides crowdsourcing annotations from multiple levels of expertise with three types of labeling confidence.

\begin{table}[t]
\begin{center}
\begin{tiny}
\begin{sc}
\caption{Comparison of datasets containing multiple annotations per sample and multiple levels of expertise in annotations.}
\begin{tabular}
{p{0.12\textwidth}p{0.12\textwidth}p{0.12\textwidth}p{0.1\textwidth}p{0.1\textwidth}p{0.08\textwidth}p{0.04\textwidth}p{0.04\textwidth}p{0.15\textwidth}}
\toprule
\textbf{Dataset} & \textbf{Expertise} & \textbf{Annotations per Sample} & \textbf{Annotation Confidence} & \textbf{Resolution} & \textbf{Samples} & \textbf{Classes} &  \textbf{Total Annotations} \\ 
\toprule
CIFAR10-H \cite{peterson2019human}       & Novices (2,571)       &  50    &   \ding{55}   &       32x32       &       10,000       &       10       &       500k       \\ 
CIFAR-10S \cite{collins2022eliciting}       & Novices (248)              &       6       &       \ding{51} (3)     &       32x32       &       1,000       &       10       &       6,200       \\ 
\midrule
LIDC-IDRI \cite{armato2011lung}       & Experts (12 radiologists)       &       4       &       \ding{55}       &       512x512       &       1018       &       2       &       4,072       \\ 
MICCAI2012 \cite{litjens2012pattern}       & Experts (observers)                          &            3                               &          \ding{55}                  &              320x320               &  48 &  2 & 144\\ 
RIGA \cite{almazroa2017agreement}       & Experts (6 ophthalmologists)       &       6       &    \ding{55}                        &           2160x1440       &  750          &  2 & 4500\\ 
QUBIQ \cite{menzequantification}       & Experts&       4       &       \ding{55}       &       N/A (>640x640)       &       268       &       3       &       1049       \\ \midrule
Phrase Detectives Corpus V2 \cite{poesio2019crowdsourced}       &  Novices (1958), Experts (2)       &       20       &       \ding{55}       &       N/A       &       108,000       &       3       &       2.2M       \\ \midrule
         % &                       &                                         &                            &                            &  &  \\ \hline
\texttt{CRACKS}         & Novices (26), practitioners (8), expert (1)                     & 35                                        &    \ding{51}  (3)                       &       701x255                     &  7636  & 3 &  231729  \\ \bottomrule
\end{tabular}
\label{tab:stats_datasets}
\end{sc}
\end{tiny}
\end{center}
\end{table}

\begin{table}[t]
\begin{center}
\begin{sc}
\begin{tiny}
\caption{Statistics of the available data modalities in the train and test sets.}
\begin{tabular}{lllllllll}
\toprule
\multicolumn{1}{c}{\multirow{3}{*}{Expertise}} & \multicolumn{4}{c}{Train Total} & \multicolumn{4}{c}{Test Total} \\ %$& \multicolumn{1}{c}{\multirow{2}{*}{Crowdsourcing statistics}} \\
\multicolumn{1}{c}{}                           & \multicolumn{2}{c}{Bounding boxes}       & \multicolumn{2}{c}{percentage of pixels(\%)}      & \multicolumn{2}{c}{Bounding boxes}      & \multicolumn{2}{c}{percentage of pixels(\%)}                                                \\ \cline{2-5} \cline{6-9}
\multicolumn{1}{c}{}                           & {Certain}       & {Uncertain}      &  {Certain}      & {Uncertain}      & {Certain}       & {Uncertain}      &  {Certain}      & {Uncertain} \\ \midrule
Geophysicist                                   &       118       &        697          &        0.4844     &        2.0554            & 732     & 6089            & 0.3416     & 2.1265  \\
Practitioners                                  &              11376 &         36688     &              0.6480 &         1.6992       &       ---      &       ---        &       ---      &       --- \\
Novices                                        &      84650        &       91379           &      1.6762        &       1.9238           &     ---        &     ---          &     ---        &     ---     \\ \bottomrule
\end{tabular}
\label{tab:stats_data}
\end{tiny}
\end{sc}
\end{center}
\end{table}

\begin{figure}[t]
\centering
\includegraphics[scale =.4]{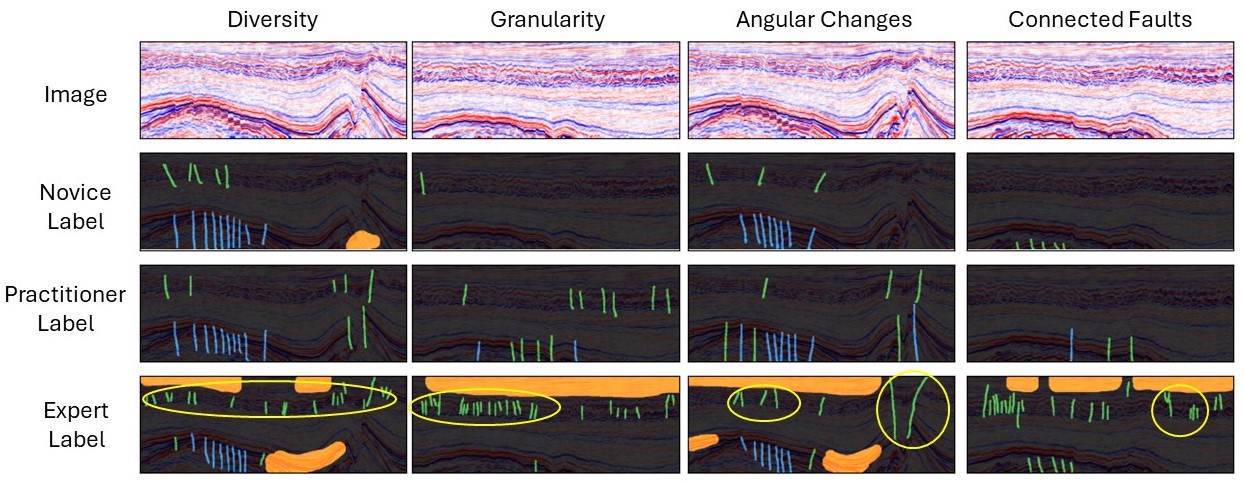}
\vspace{-1.5mm}
\caption{Expert/Practitioner/Novice annotation of seismic sections.}\label{fig:challenges}\vspace{-3mm}
\end{figure}
\vspace{-3mm}
\section{Dataset Details}
\label{sec:Dataset}
\vspace{-3mm}
Table~\ref{tab:stats_datasets} shows the statistics of \texttt{CRACKS} dataset. Each section from the F3 block is an image of resolution $255 \times 701$ pixels with $255$ pixels placed along the depth from the surface of the earth. There are $400$ such sections, with an average of $19$ faults in each section. More details regarding the geological model is available in Appendix~\ref{subsec:F3}. 
\vspace{-3mm}
\paragraph{Amazon Mechanical Turk Setup} We leverage the Amazon Mechanical Turk (MTurk) platform to crowdsource fault annotations, adapting existing image segmentation layouts to suit our needs. In particular, we divide the 400-section F3 dataset into 20 batches consisting of 20 Human Intelligence Tasks (HITs), each batch corresponding to a single section. The data collection process was deemed to be Minimal risk research qualified for IRB exemption status under 45 CFR 46 104d.2 (Protocol H23360). All annotators are between 18 – 89 years old, and are able to consent without requiring surrogate consent from a legally authorized representative. Screenshots from the MTurk setup as well as other details are provided in Appendix~\ref{subsec:MTurk_setup}.
\vspace{-3mm}
\paragraph{Annotation Expertise} Annotations in \texttt{CRACKS} dataset are sourced from 26 novices, 8 practitioners, and a geophysicist. The novices, practitioners and geophysicist label on the same seismic sections from the F3 block. Hence, there are $35$ annotations per fault (Table~\ref{tab:stats_datasets}, column 3). Our definition of expertise is as follows:\vspace{-3mm}
\begin{itemize}
    \item \textbf{Novices} Recruits from MTurk with no experience interacting with subsurface sections and faults. To guide the labeling task, the novices are shown a \href{https://www.youtube.com/watch?v=Fyr4b_RiT8s}{5 min instructional video} that introduces faults and fault annotation on the MTurk platform. More information regarding worker compensation, labeling instructions, and eligibility is present on our \href{https://alregib.ece.gatech.edu/fun-ml-fault-uncertainty-for-machine-learning/}{participation call webpage} as well as Appendix~\ref{sec:AppC}.
    \item \textbf{Practitioners} Practitioners are graduate students who have previously worked with seismic data. However, they have not specifically labeled faults. The same \href{https://alregib.ece.gatech.edu/fun-ml-fault-uncertainty-for-machine-learning/}{participation call webpage} applies to practitioners, except participation is limited to Georgia Tech students.
    \item \textbf{Expert} A geophysicist acted as the labeling expert. The sixth column in Table~\ref{tab:stats_datasets} with $7636$ samples in \texttt{CRACKS} refer to the number of faults annotated by the expert. 
\end{itemize}

\begin{figure}[t]
\centering
\includegraphics[scale =.3]{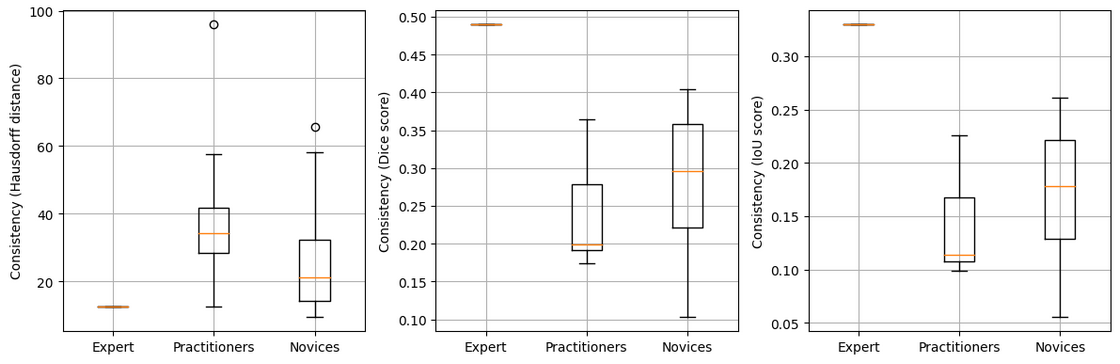}
\vspace{-1.5mm}
\caption{Consistency metrics (section-wise) across $120$ repeated sections.}\label{fig:consistency}\vspace{-3mm}
\end{figure}

\paragraph{Annotation Confidence Types} Four exemplar seismic sections along with randomly selected novice, practitioner, and the expert labels are shown in Fig.~\ref{fig:challenges}. The color of the labels (blue, green, or orange) depicts the confidence of the annotators for that fault. Confidence definitions are as follows:
\begin{itemize}
    \item \textbf{No fault (certain)}: This category is used to demarcate regions where the annotator is sure of the absence of any seismic faults. This region is annotated in orange in Fig.~\ref{fig:challenges}.
    \item\textbf{Fault (certain)}: Used to delineate seismic faults of whose existence the annotator feels fairly confident in. We use the color blue for this category in Fig.~\ref{fig:challenges}. 
    \item \textbf{Fault (uncertain)}: Used to delineate faults that are only partially discernible or are otherwise not fully ascertainable by the annotator. We use the color green for this category in Fig.~\ref{fig:challenges}. 
\end{itemize}
The specifics of the brush tool used to obtain these annotations on MTurk platform are provided in Appendix~\ref{subsec:MTurk_setup}. In Table~\ref{tab:stats_data}, we show the number of certain and uncertain faults labeled by the expert geophysicist and aggregates of all $26$ novices and $8$ practitioners. Interestingly, of the $7636$ faults labeled by the expert, only $815$ faults are certain. This is only $14.5\%$ of all faults labeled by the geophysicist. In contrast, $48.1\%$ of all faults labeled by the novices are certain. However, in accordance with our granted IRB exemption, we do not collect any behavioral data that may lead to analyzing this confidence inequity. Additionally, averaged percentage of pixels in each section describes the class-imbalance issues in section-wise segmentation.
\vspace{-3mm}
\paragraph{Annotation Quality Control} To ensure quality of obtained labels, each batch of $20$ HITs is augmented twice with 2 redundant and randomly selected seismic sections from the batch. Hence, every batch has $24$ sections, out of which only $20$ are unique. None of the annotators across expertise are told this in order to perform a quality assurance (QA) analysis. Hence, we obtain $120$ quality assessment (QA) sections ($40$ sections repeated thrice) within the dataset to evaluate annotator self-consistency. We use three metrics - modified Hausdorff distance, mean Intersection over Union (mIOU), and DICE coefficient to measure this self-consistency. The specifics of the metrics are presented in Appendix~\ref{subsec:Metrics}. Hausdorff distance is sensitive to the boundary of the faults while mIOU and DICE coefficients are sensitive to the areas of the fault. All three metrics are commonly used to quantify fault delineation~\cite{dou2021attention} as well as medical segmentation~\cite{taha2015metrics}. The results from these metrics for the expert as well as averaged novice and practitioners are showcased in Fig.~\ref{fig:consistency}. Note that the expert outperforms the novices and practitioners on all self-consistency metrics. The self-consistency metrics showcases precision of annotators. However, labeling bias may still exist, which is expanded on in the Limitations section~\ref{subsec:limitations}. Additional results and statistics including label category usage, labeling time, confidence metrics, and intra- and inter-annotator agreements are provided in Appendix~\ref{subsec:Statistics}. To ensure label quality, all annotators were compensated with bonus pay based on the self-consistency QA metrics. Also, to ensure novice retention across all $480$ sections, the annotators were paid on a prorated pay, in which the final 160 section annotations are worth increasingly more for every 10 sections annotated. Further details regarding payscales are presented in Appendix~\ref{subsec:MTurk_compensation}. 
\paragraph{Annotation Challenges} We present annotations from practitioners, novices, and the expert in Fig.~\ref{fig:challenges}. Fig.~\ref{fig:challenges} showcases (i) the challenges within the dataset, and (ii) expert knowledge that allows overcoming these challenges. The main challenges are diversity, granularity, angular changes, and connected faults. Diversity refers to the wide variety of faults at different locations within the volume. Granularity refers to faults that are harder to distinguish due to a inconsistent shape or blending with surrounding lithographic reflections. Angular changes refers to faults that branch out in multiple directions. Connected faults refer to faults that intersect with each other. Note that the existence of connected faults motivates the need to include both semantic segmentation and instance segmentation tasks. From a semantic segmentation perspective, a connected fault can be viewed as a single entity, but from an instance segmentation perspective, they are separate instances. Additionally, visualizing these challenges, provides insight into the knowledge requirements of experts. For example, by looking at the diversity and granularity columns, we see that the expert is better able to identify faults in diverse regions of the seismic section as well as more difficult to identify faults compared to both the practitioner and novice. Furthermore, the expert is more proficient at capturing complex fault geometries such as angles and connections between faults compared to non-experts. Additionally, there is a qualitative improvement in annotations when going from novices to practitioner labels. This demonstrates the need to understand label disagreement at multiple expertise levels. 

\section{Experiments}
\label{sec:Experiments}
In this section, we demonstrate the following on the bounding box annotations from Table~\ref{tab:stats_data}:
\begin{itemize}
    \item In Section~\ref{subsec:Model-free}, we provide a sanity check to showcase that the obtained annotations from novices and practitioners are useful for fault delineation.
    \item In Section~\ref{subsec:Detect}, we showcase the value of adding novice and practitioner labels to existing ML workflows for the applications of fault detection and fault instance segmentation.
    \item In Section~\ref{subsec:SSL}, we demonstrate the utility of \texttt{CRACKS} dataset for self-supervised learning under the presence of large, albeit noisy, labels.
    \item Finally, in Section~\ref{subsec:Semantic}, we show that section-wise segmentation using state-of-the-art segmentation models fail on imbalanced \texttt{CRACKS} data.
\end{itemize}

The detailed statistics of the dataset that are used for model training and evaluation is shown in Table \ref{tab:stats_data}. Note that inference is always evaluated against expert labels. The goal is to utilize the available novice and practitioner labels to approximate expert annotations, all on the same seismic images. In this paper, we do not utilize the orange \emph{No Fault (certain)} labels. Also, no differentiation is made between certain and uncertain labels. Both sets of labels are treated as faults. In Appendix~\ref{sec:AppE}, we provide additional results that fuse between certain and uncertain labels. Also, in accordance with fault delineation literature~\cite{dou2021attention, mustafa2024visual} around $10\%$ of the expert ground truth ($815$ faults out of $7636$ from unique seismic images) is used during training. Hence, the goal is to adapt the novice and practitioner labels to the expert label domain. This is because, in practical settings, very large volumes of the earth can be surveyed but annotating them via experts in not feasible. We aim to show that crowdsourcing annotations on these volumes
is a viable strategy to analyze subsurfaces.
\begin{figure}[t]
\centering
\includegraphics[scale =.3]{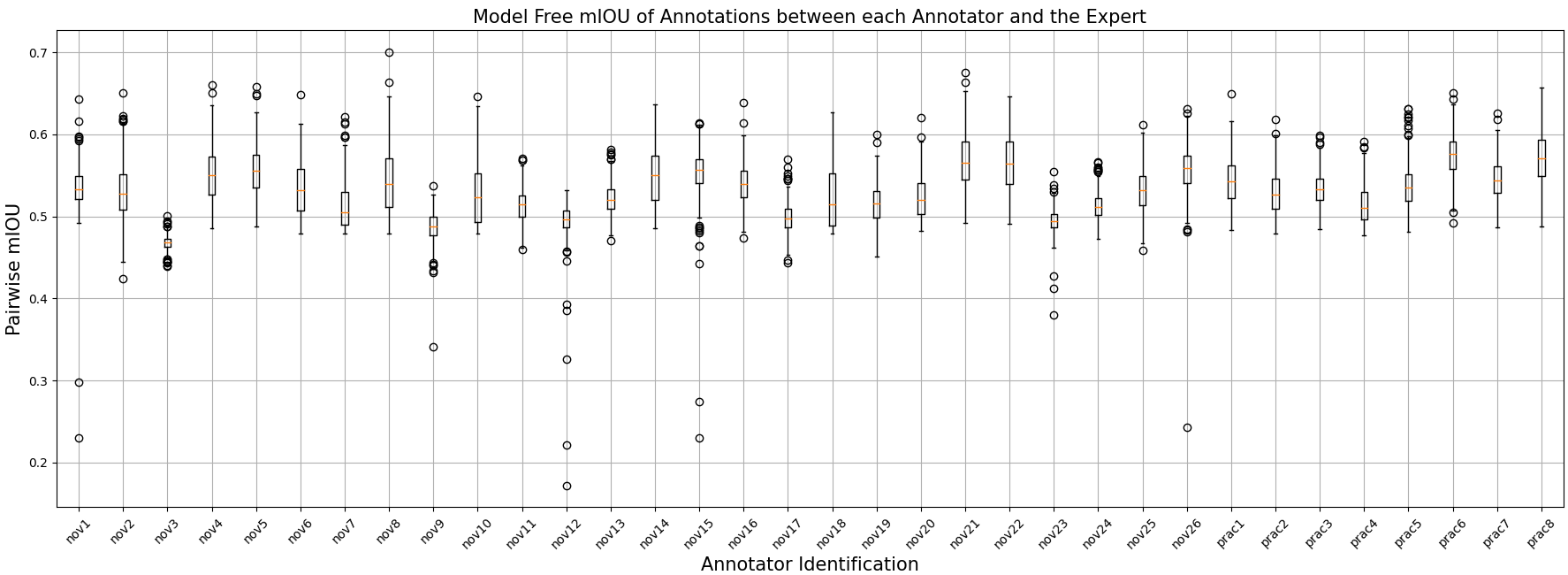}
\vspace{-1.5mm}
\caption{Sanity check to demonstrate that the novice and practitioner annotations are not trivial. Pairwise mIOU between annotations from every practitioner and novice (\texttt{x-axis}) is calculated against the expert annotation and is presented on the \texttt{y-axis}.}\label{fig:model_free}\vspace{-3mm}
\end{figure}

\subsection{Model-free Results}
\label{subsec:Model-free}
Since the same annotations occur across the same seismic sections and faults, we run the mIOU, DICE, and Hausdorff distance metrics between all sections. In Figure \ref{fig:model_free}, we show the distribution of mIOU scores calculated between the annotated seismic sections of each annotator and the expert annotator. We show the distribution of mIOU scores across all sections with an associated box and whisker plot for each annotator. The scores indicate that the novices and practitioners are not producing trivial annotations and each are acting to label an approximation of the true fault distribution of the dataset. Hence, all the novice and practitioner annotations can be considered as noisy labels of the true expert ground truths. Recent works in natural images~\cite{wei2021learning} and medical images~\cite{ju2022improving} tackle this noisy label phenomenon. However, as demonstrated in Fig.~\ref{fig:challenges}, the disagreements among novices and practitioners, and between them, is considerably large. This is because of the knowledge gap between experts and practitioners, and practitioners and novices.

\begin{table}[t]
\begin{center}
\begin{tiny}
\begin{sc}
\caption{Results of fault detection and instance segmentation for different training sets and methods using mean Average Precision at 0.5. The best results in each application are bolded.}
\begin{tabular}{lcr}
\toprule
Training Set  & \multicolumn{1}{c}{Detection} & Instance Segmentation  \\
\midrule

Expert     & 0.425 ± 0.004   &      0.145 ± 0.002      \\
\midrule
Novice (Average)     & 0.219 ± 0.001   &  0.051 ± 0.001            \\
Practitioner (Average)     & 0.278 ± 0.001     &  0.074 ± 0.001       \\
\midrule
Novice expert fine-tuned (Average)     & 0.460 ± 0.001   &    0.146 ± 0.001        \\
Practitioner expert fine-tuned (Average)      & \textbf{0.466 ± 0.001}  &  0.152 ± 0.001            \\ 
\midrule
Practitioner+Novice (Label Fusion)     & 0.354 ± 0.004      &   0.103 ± 0.002      \\
Practitioner+Novice (Label Fusion) expert fine-tuned      & 0.457 ± 0.003      &  \textbf{0.177 ± 0.002}         \\
\bottomrule
\end{tabular}
\vspace*{.1mm}
\label{yolodetectionresults}
\end{sc}
\end{tiny}
\end{center}
\end{table}

\subsection{Detection and Instance Segmentation}
\label{subsec:Detect}
Fault detection is the first step of the fault delineation process. For the detection task, we convert the segmentation masks provided by annotators into surrounding bounding boxes. We then train a YOLOv5-l \cite{glenn_jocher_2022_7347926} object detector for each annotator's label set. All models are trained for 100 epochs with an image size of 640 pixels.  For the expert training set, we train with $815$ faults from 40 sections; for novices and practitioners, we use all available faults across 400 sections. We evaluate all models using $6821$ faults from the expert (Table~\ref{tab:stats_data}). Additionally, we fuse the labels from practitioners and novices to create a new training set. Simple classification experiments have benefited from majority voting~\cite{sheng2008get} from MTurk data. In our fusion strategy (rows 6 and 7 in Table~\ref{yolodetectionresults}), we give practitioner labels twice the weight of novice labels. We also perform experiments where expert data is used to fine-tune models previously trained on novice or practitioner data for both the segmentation and detection tasks. We display all results for the expert test set in Table \ref{yolodetectionresults}. 

Next, we perform fault instance segmentation where the goal is to segment specific instances of different faults present in the seismic volume. For the instance segmentation task, we convert the segmentation masks provided by annotators into polygon annotations. We then train a YOLOv5l-seg \cite{glenn_jocher_2022_7347926} instance segmentation model for each annotator's label set. We follow the same procedure as detection for all experiments. We display all results for the expert test set in Table \ref{yolodetectionresults}. For both detection and instance segmentation, we observe that models fine-tuned with expert data show superior performance compared to those trained solely on expert data. This demonstrates that incorporating labels from both practitioners and novices significantly enhances model performance for fault detection compared to using exclusively the expert's training data.

\begin{figure}[t]
\centering
\includegraphics[scale =.3]{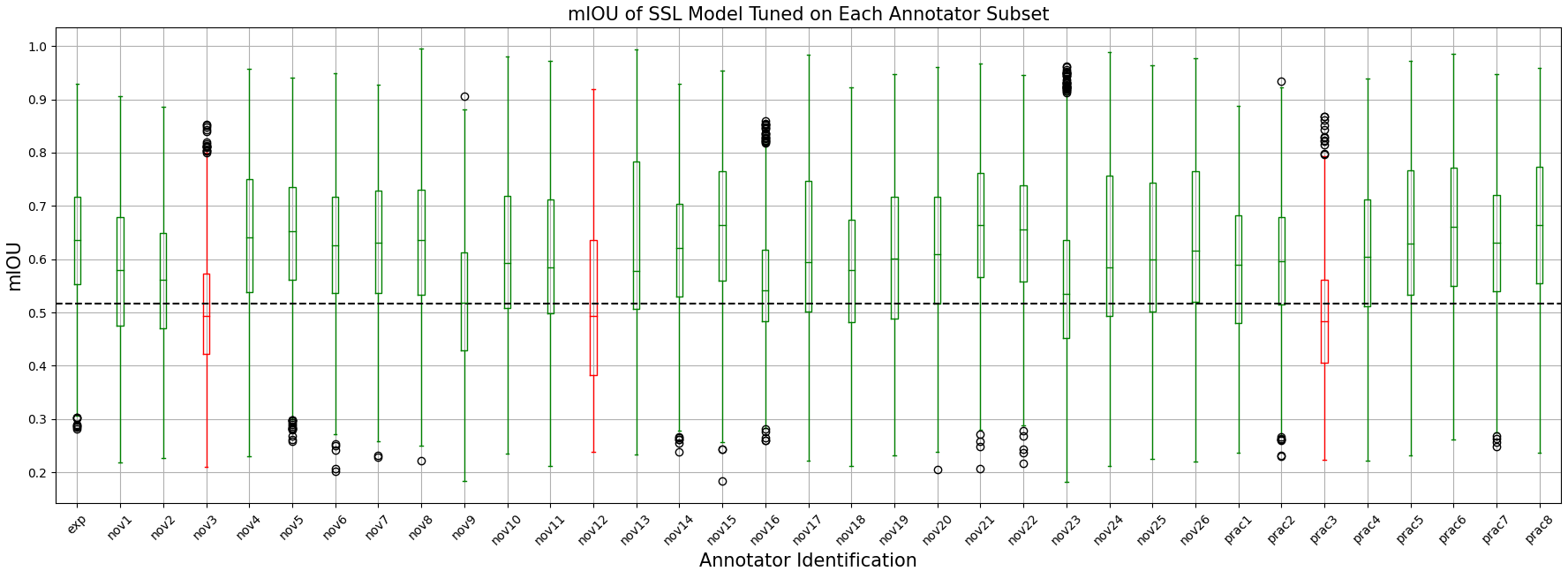}
\vspace{-1.5mm}
\caption{SSL finetuning on each annotator subset.}\label{fig:ssl}\vspace{-3mm}
\end{figure}

\subsection{Self-supervised Experiments}
\label{subsec:SSL}
In Figure \ref{fig:ssl}, we demonstrate the potential of \texttt{CRACKS} to serve as a new type of benchmark within the context of self supervised learning. To do this, we take the original fault labeled dataset on the semantic segmentation task for each annotator and extract out the cropped region where the annotation occurs. This forms a dataset consisting of object crops where each crop represents where the annotator believes a fault exists. We train a ResNet18 \cite{he2016deep} with the SimCLR \cite{chen2020simple} contrastive learning algorithm for 100 epochs on the overall pool of fault region crops. The training uses a stochastic gradient optimizer (SGD), .001 learning rate, and the default augmentation policy as the original paper. Next, we attach a segmentation head from the UNet \cite{ronneberger2015u} architecture and freeze the weights of the backbone ResNet18 network. We then fine-tune the segmentation head on each individual annotator labeled object crop dataset for 50 epochs with a binary cross-entropy loss, SGD optimizer, and a learning rate of .001. Each model is tested on the same held out expert testing fault object crop segmentation data. The distribution of the IOU scores for each annotator on the expert test set are visually shown as a box and whisper plot in Figure \ref{fig:ssl}. We highlight all annotators in green where the median of the box and whisper plot of their IOU scores is above the overall average of model free comparisons with the expert from Figure \ref{fig:model_free}. We note that leveraging SSL algorithms can noticeably improve fault segmentation performance despite the difficulty of the associated task. 

\begin{table}[h!]
\begin{center}
\begin{scriptsize}
\begin{sc}
\caption{Averaged novice and practitioner Section-wise Fault Segmentation Results before and after finetuning on expert. Hausdorff distance shows significant decrease (in red) while all other changes are statistically insignificant.}
\begin{tabular}{lcccccr}
\toprule
Models & \multicolumn{3}{c}{Before Finetuning} & \multicolumn{3}{c}{After finetuning}\\ 
& mIOU & Hausdorff & DICE Score & mIOU & Hausdorff & DICE Score \\
\midrule
Segformer \cite{xie2021segformer}                          & 0.53  & 41.7  & 0.58 & 0.54 & \color{red}{59.1} & 0.56  \\
Unet \cite{ronneberger2015u}                            & 0.48  & 74.5  & 0.49 & 0.48 & \color{red}{87.6} & 0.5  \\
Unet++ \cite{zhou2019unetplusplus}                     & 0.49  & 75.53 & 0.5 & 0.49 & \color{red}{98.9} & 0.51  \\
Deeplab v3 (Resenet 50 \cite{he2016deep}) \cite{chen2017deeplab}   & 0.48  & 54.7  & 0.53 & 0.48 & \color{red}{122.1} & 0.51 \\
\bottomrule
\end{tabular}
\vspace*{.1mm}
\label{table:segmentation_results_beforefine}
\end{sc}
\end{scriptsize}
\end{center}
\end{table}

\subsection{Section-wise Segmentation}
\label{subsec:Semantic}
For the segmentation task, we evaluate the dataset using State-Of-The-Art (SOTA) segmentation models that includes SegFormer \cite{xie2021segformer}, UNet \cite{ronneberger2015u}, UNet++ \cite{zhou2018unet++}, and DeepLab v3 \cite{chen2017rethinking} (ResNet-50 \cite{he2016deep} as the backbone). Training choices include the cross-entropy loss, Adam optimizer \cite{kingma2017adam}, and 100 epochs of training. We also consider two types of training settings: novice training on top of the pre-trained SOTA network and fine-tuning these trained models with additional expert data. We refer to these settings as before and after fine-tuning respectively. These results are shown in Table \ref{table:segmentation_results_beforefine}. We note poor performance scores across all SOTA models. Additionally, further expert fine-tuning worsens Hausdorff distance metric or causes statistically insignificant changes. This, along with the results of Figure \ref{fig:ssl} showcases the difficulty of the fault segmentation task, which motivates further advancements in segmentation training paradigms to address the challenges of this dataset as discussed previously in Figure \ref{fig:challenges}. Additional results are presented in Appendix~\ref{sec:AppE}.

\section{Discussion and Conclusion}

\subsection{Limitations}
\label{subsec:limitations}
\paragraph{Labeling bias} Geophysicists who label faults in seismic sections generally do so based on applications they are interested in. For instance, geophysicists monitoring earthquakes are only interested in active faults that cause rocks to move laterally. Geophysicists looking for carbon sequestration are only interested in sealing faults that do not allow movements between rocks. Hence, datasets constructed on seismic sections of interest might have differently labeled faults based on the geophysicist labeling it~\cite{mustafa2024visual}. In practice, the time-consuming nature of segmenting each fault as well as the vast quantity of subsurface data in the earth, precludes having multiple experts looking at the same faults. In this paper, we provide consistency of labeling the same frames across time as quality control to ensure that the expert is precise. However, bias may exist, depending on specific task requirements.

\paragraph{2D vs 3D labeling} In this paper, we choose to annotate faults on 2D sections, even though the F3 dataset is available in 3D. We do so because of two reasons. Firstly, it is not trivial to extend MTurk software to work on 3D data. Existing methods that annotate 3D images treat them as 2D data which are postprocessed~\cite{giuly2013dp2}. This is an additional workflow, which is not the aim of the paper. Moreover, evaluation of 3D segmented faults is done in 2D~\cite{li2022automatic}. However, the $400$ annotated sections in our dataset are contiguous frames that make up the volume and further methods can be developed that treat fault annotations as 3D data.

\paragraph{Visual labeling only} During the annotation process, the expert was instructed to only utilize visual seismic data and work on MTurk software, to not bias the expert labels against the novices and practitioners. However, in practice, geophysicists utilize specialized software like OpendTect~\cite{dgbpy_2019} and use multi-modal data including well logs, rock properties, and elastic impedance among others to segment difficult faults. In this paper, we have specifically chosen F3 subsurface data to annotate faults since the faults in the Netherlands North Sea are easier to visually inspect and label.

\subsection{Conclusion} 
The \texttt{CRACKS} dataset presents the first subsurface fault-labeled data on the F3 seismic volume using resources from crowdsourcing platforms. \texttt{CRACKS} allows the study of annotation disagreement within and between expertise levels on the non-common knowledge task of seismic fault delineation. The experimental design provides real-world case studies where the underlying images in the train and test sets are the same. However, the labels in the crowdsourced data can be considered as noisy versions of the true ground truth labels, thereby significantly challenging existing ML paradigms. This is a common occurrence in medical data and even more in seismic images where the vast regions in the earth's subsurface are not surveyed and hence, the physical properties unknown. \texttt{CRACKS} provides the first foray into analyzing and delineating subsurface faults based on novices from crowdsourcing platforms.

\begin{ack}
This work is supported by the ML4Seismic Industry Partners
at Georgia Tech.
\end{ack}

\bibliographystyle{IEEEbib}
\bibliography{references}

\newpage

%%%%%%%%%%%%%%%%%%%%%%%%%%%%%%%%%%%%%%%%%%%%%%%%%%%%%%%%%%%%
\section*{Checklist}

\begin{enumerate}

\item For all authors...
\begin{enumerate}
  \item Do the main claims made in the abstract and introduction accurately reflect the paper's contributions and scope?
    \answerYes{}
  \item Did you describe the limitations of your work?
    \answerYes{See section~\ref{subsec:limitations}.}
  \item Did you discuss any potential negative societal impacts of your work?
    \answerNo{While we expand on the multiple limitations in the dataset, we stand by its propensity for positive societal impact.}
  \item Have you read the ethics review guidelines and ensured that your paper conforms to them?
    \answerYes{}
\end{enumerate}

\item If you are including theoretical results...
\begin{enumerate}
  \item Did you state the full set of assumptions of all theoretical results?
    \answerNA{}
	\item Did you include complete proofs of all theoretical results?
    \answerNA{}
\end{enumerate}

\item If you ran experiments (e.g. for benchmarks)...
\begin{enumerate}
  \item Did you include the code, data, and instructions needed to reproduce the main experimental results (either in the supplemental material or as a URL)?
    \answerYes{}
  \item Did you specify all the training details (e.g., data splits, hyperparameters, how they were chosen)?
    \answerYes{}
	\item Did you report error bars (e.g., with respect to the random seed after running experiments multiple times)?
    \answerYes{}
	\item Did you include the total amount of compute and the type of resources used (e.g., type of GPUs, internal cluster, or cloud provider)?
    \answerYes{}
\end{enumerate}

\item If you are using existing assets (e.g., code, data, models) or curating/releasing new assets...
\begin{enumerate}
  \item If your work uses existing assets, did you cite the creators?
    \answerYes{}
  \item Did you mention the license of the assets?
    \answerYes{}
  \item Did you include any new assets either in the supplemental material or as a URL?
    \answerYes{}
  \item Did you discuss whether and how consent was obtained from people whose data you're using/curating?
    \answerYes{}
  \item Did you discuss whether the data you are using/curating contains personally identifiable information or offensive content?
    \answerYes{No personally identifiable information is collected or distributed.}
\end{enumerate}

\item If you used crowdsourcing or conducted research with human subjects...
\begin{enumerate}
  \item Did you include the full text of instructions given to participants and screenshots, if applicable?
    \answerYes{}
  \item Did you describe any potential participant risks, with links to Institutional Review Board (IRB) approvals, if applicable?
    \answerYes{}
  \item Did you include the estimated hourly wage paid to participants and the total amount spent on participant compensation?
    \answerYes{}
\end{enumerate}

\end{enumerate}

%%%%%%%%%%%%%%%%%%%%%%%%%%%%%%%%%%%%%%%%%%%%%%%%%%%%%%%%%%%%

\appendix

\newpage

\section{Dataset, Participation call, and Benchmarks access}
\label{sec:AppA}

\subsection{Links to Access Dataset}
\label{app: links}
We provide open access to the dataset. The images and labels found in the CRACKS dataset are present at:\\
 \href{https://zenodo.org/records/11559387}{Dataset Access}
 
The benchmarks provided in the paper are accessible at the following link:  \\
 \href{https://github.com/olivesgatech/CRACKS}{Code Access} 

\subsection{Licenses and DOI}
\label{app: license}
The code is associated with an MIT License. The DOI of the dataset is 10.5281/zenodo.11559386. The associated license with the dataset is a Creative Commons International 4 license.

\subsection{Maintenance Plan}
\label{app: maintenance}

The code will be hosted within the github repository specified in Section \ref{app: links}. Instructions and details regarding the dataset will be located at this same repository. Images and labels for the dataset are located at the zenodo directory in Section \ref{app: links}. Labels for these images will be included within this same zenodo dataset after acceptance of the paper. Additional data from other expert annotators will be added over time as part of our partnership with various contacts in the geophysics community. Within the Github repository, we will maintain a comprehensive survey of all literature that use the CRACKS dataset. This will include a unified result table and access to publicly available github repositories that benchmark on CRACKS. 

\subsection{Dataset Folder Structure}
\label{app: structure}

\paragraph{Images} The images for the full 400 images corresponding to each seismic section are placed in a folder called images.zip. Every image is named with a convention that denotes the position within the 3D seismic volume that the section is drawn from.

\paragraph{Labels}
The labels exist within a separate folder called Fault Segmentations.zip. This folder contains 35 directories that each correspond to the annotations associated with our 26 novices, 8 practitioners, and the single expert. The directories are named in an intuitive manner to indicate which annotator created the associated labels. Examples include novice01, practitioner2, and expert. Within each folder named by the associated annotator are the fault annotations for each seismic section that that specific annotator worked on. The label files exist in a .png format with a naming convention that indicates which seismic section from the overall volume that these labels correspond to. As discussed in the main paper, every label file has three colors that indicates confident existence of fault (blue), uncertain existence of fault (green), and confidence of the non-existence of a fault. 

\subsection{Reproducibility Statement and Attributions}
\label{app: attributions}

We make partial use of the codes provided below:

\href{https://github.com/HobbitLong/SupContrast}{SimCLR } \\
\href{https://github.com/olivesgatech/Image-2022-Volumetric-Contrastive}{Volumetric Contrast} \\ 
\href{https://github.com/facebookresearch/vicreg}{VicReg} \\ 

We also compare against SOTA supervised segmentation methods making use of code from the following repository:

\href{https://github.com/qubvel/segmentation_models.pytorch}{Segmentation PyTorch}\\

Additionally, object detection functionality from the from the following Yolo v5 repository is used:

\href{https://github.com/ultralytics/yolov5}{Yolo v5}\\

We also show a comparison in Section \ref{sec:AppE} with synthetically trained models using the following codebase: 

\href{https://github.com/xinwucwp/faultSeg}{FaultSeg3D}\\

Results for our paper can be replicated using the code, images, and labels found in Section \ref{app: links}.

\begin{figure}[t]
\centering
\includegraphics[scale =.3]{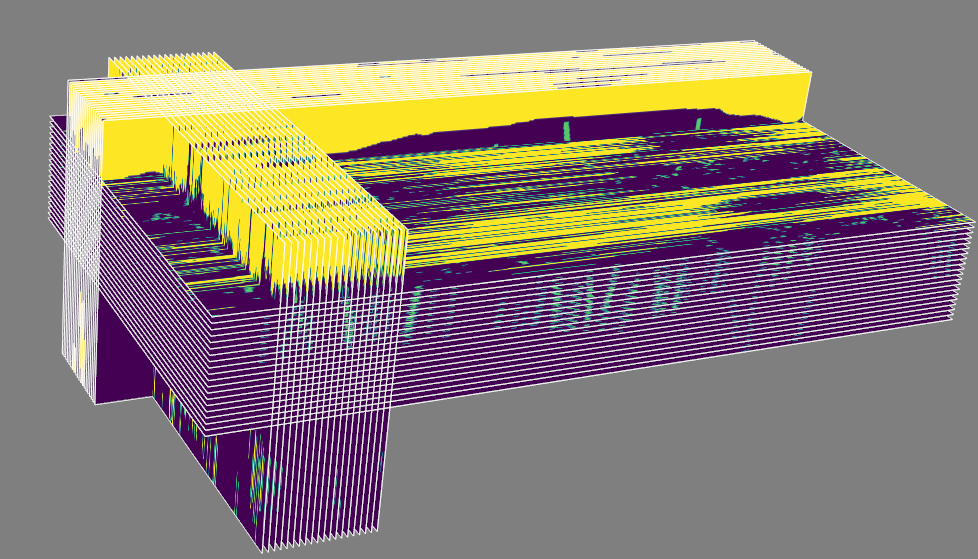}
\vspace{-1.5mm}
\caption{3D visualization of fault labels across entire volume.}\label{fig:3D}\vspace{-3mm}
\end{figure}

\section{Dataset Details}
\label{sec:AppB}
\subsection{F3 Block}
\label{subsec:F3}
Seismic datasets are obtained by surveying different subsurface regions in the Earth. For instance, the popular F3 dataset~\cite{F3_data} is a rectangular volume of dimensions $16\text{ km}\times24\text{ km}$ located off the shores of the Netherlands North Sea. This dataset is divided into 400 sections, with each section having a resolution of $255\times701$ pixels. Hence, each pixel roughly corresponds to $25\text{ m}\times25\text{ m}$ of the earth's subsurface. Note that the 400 sections combine to produce a 3D volume. This volume along with the expert fault labels are shown in Fig.~\ref{fig:3D}. The faults are visible in green.

\subsection{Ground Truth}
\textit{Ground truth} in seismic data is derived by drilling within the Earth's subsurface. Rocks, and sediments are collected and their properties are analyzed before matching these properties to the seimsic data. Note that this type of Ground Truth is impossible to obtain across the subsurface. Instead, geophysicists construct Earth models based on wave reconstructions of known properties, such as velocity, and infer rock and sediment types, based again on known properties like elasticity among others. These models, along with seismic data are analyzed before converging on a model of the Earth's subsurface for that region. Note that this requires expertise and training and may take months of work before a seismic volume is analyzed.

\subsection{Available annotations}
A key feature in seismic analysis is the types of annotations available based on the same subsurface volume. The original F3 volume~\cite{F3_data} is provided by the Netherlands government without any ML-specific annotations to study the Earth's subsurface under the North Sea. The authors in~\cite{alaudah2018structure} extract four types of structures from this volume including faults, horizons, chaotic regions, and salt domes. The goal is to classify a given bounding box image into one of the four classes. Similarly, the authors in~\cite{alaudah2019machine} provide facies segmentation. In geology, facies are characterized by rocks or sediment deposits that exhibit similar lithological, chemical, and biological properties. For instance, in Fig.~\ref{fig:Concept}, the purple, cyan, yellow, and red regions are 4 separate facies. Depending on the application requirements, the number of facies maybe more or less fine-grained. The authors in~\cite{alaudah2019machine} annotate the F3 volume into $6$ facies. Note that these annotations are expert labeled and subjective. The subjectivity arises from the choice of the number of facies segmented.
\section{Annotation Setup}
\label{sec:AppC}
\subsection{MTurk Annotation Setup}
\label{subsec:MTurk_setup}

\begin{figure}
    \centering
    \includegraphics[width=.8\textwidth]{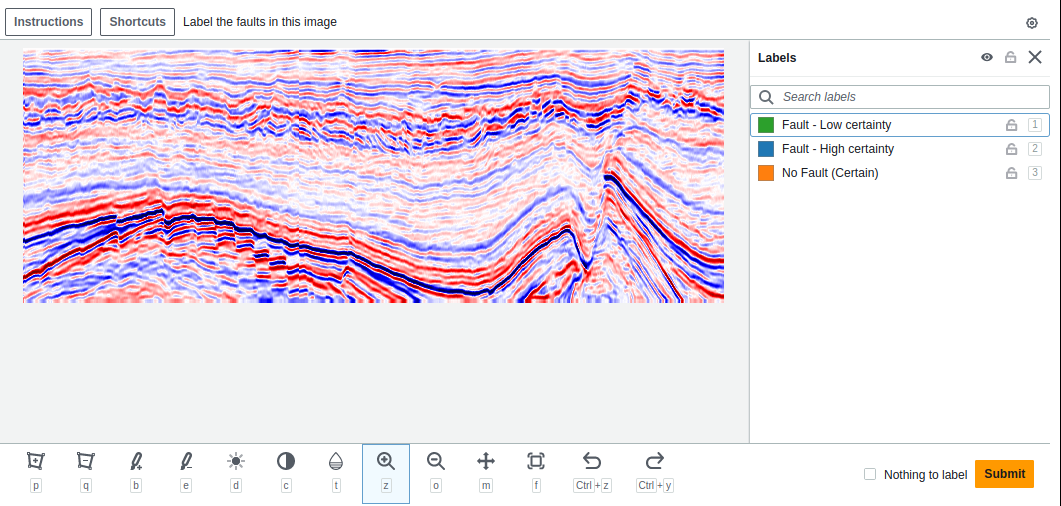}
    \caption{Fault annotation interface layout}
    \label{fig:mturk-layout}
\end{figure}

We leverage the Mturk platform for our labeling task, adapting existing image segmentation layouts to suit our needs. In particular, we divide our 400-image dataset into 20 batches consisting of 20 Human Intelligence Tasks (HITs), each corresponding to a single image. Each batch also containes 2 redundant instances of 2 randomly selected images in the batch, in order to perform a quality assurance (QA) analysis. This leads to a total of 24 images per batch seen by each annotator.
The layout of the task is depicted in Figure \ref{fig:mturk-layout}. The image is displayed on the left side of the layout, and the right toolbar provides the 3 confidence label categories for annotators to choose from: no fault (certain), fault (uncertain) and fault (certain), with colors orange, green and blue respectively. Annotators are asked to use the fault labels (green and blue) with using a small brush tool width (5 points) and the no fault label (orange) using a wider brush (20 points).

Given that a crowdsourcing platform like MTurk is highly different from a controlled laboratory setting, we need to ensure retention throughout the full annotation task as well as focus in order to ensure label quality. We achieve task completion through the use of a prorated pay bonus, in which the final 160 images are worth increasingly more for every 10 images annotated. Furthermore, we ensure focus and label quality by establishing a self-consistency bonus, in which we pay an additional amount if the annotations done for the QA images are similar, up to a threshold. Specific details on these bonuses and the full compensation are provided in the next subsection.

\subsection{MTurk Compensation}
Our goal in designing Mturk compensation metrics is to ensure \$15/hr USD wage availability to all annotators for good quality labeling and annotator retention across 400 images. We reviewed existing MTurk tasks and related literature to arrive at an initial pay rate, and then adjusted it after having team members perform the annotation task, which is approximately 4-5 hours. The final base pay rate is \$0.12 USD per image labeled, with the addition of the prorated pay and consistency bonuses. The prorated pay bonus increased the value of the last 160 labeled images, by adding \$0.01 USD to the base pay after each set of 10 images (this amounts \$71.2 USD for completing the entire dataset). The consistency bonus added \$1 USD for every batch for which the consistency score was above a given threshold, for a total of \$20 USD. Therefore, the total compensation for annotators that completed the entire labeling task is between \$71.2 and \$91.2 USD.
\label{subsec:MTurk_compensation}
\begin{figure}
    \centering
    \includegraphics[width=\textwidth]{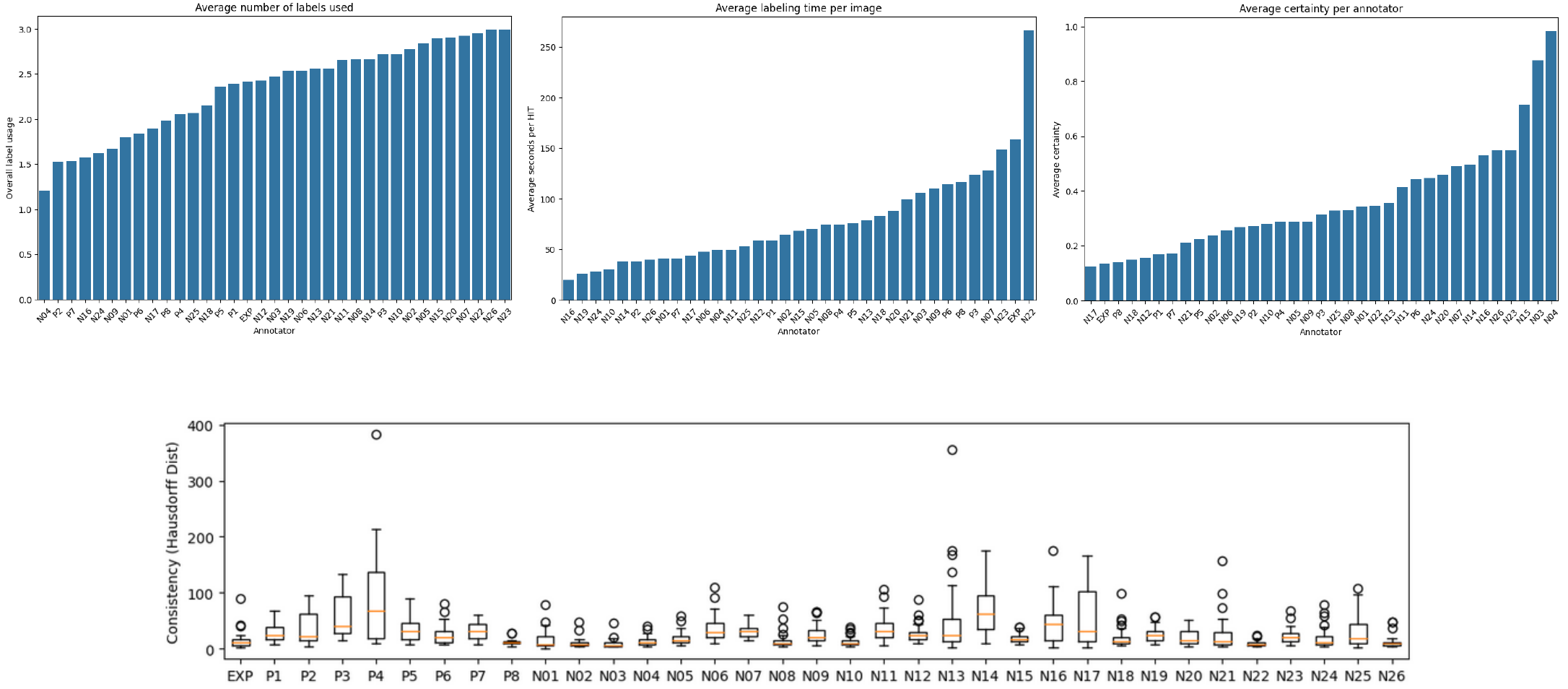}
    \caption{General dataset metrics. Top, from left to rights: average number of labels used, labeling time per image, and certainty per annotator. Bottom: individual self-consistency for each annotator on the QA images}
    \label{fig:stats1}
\end{figure}
\begin{figure}
    \centering
    \includegraphics[width=.9\textwidth]{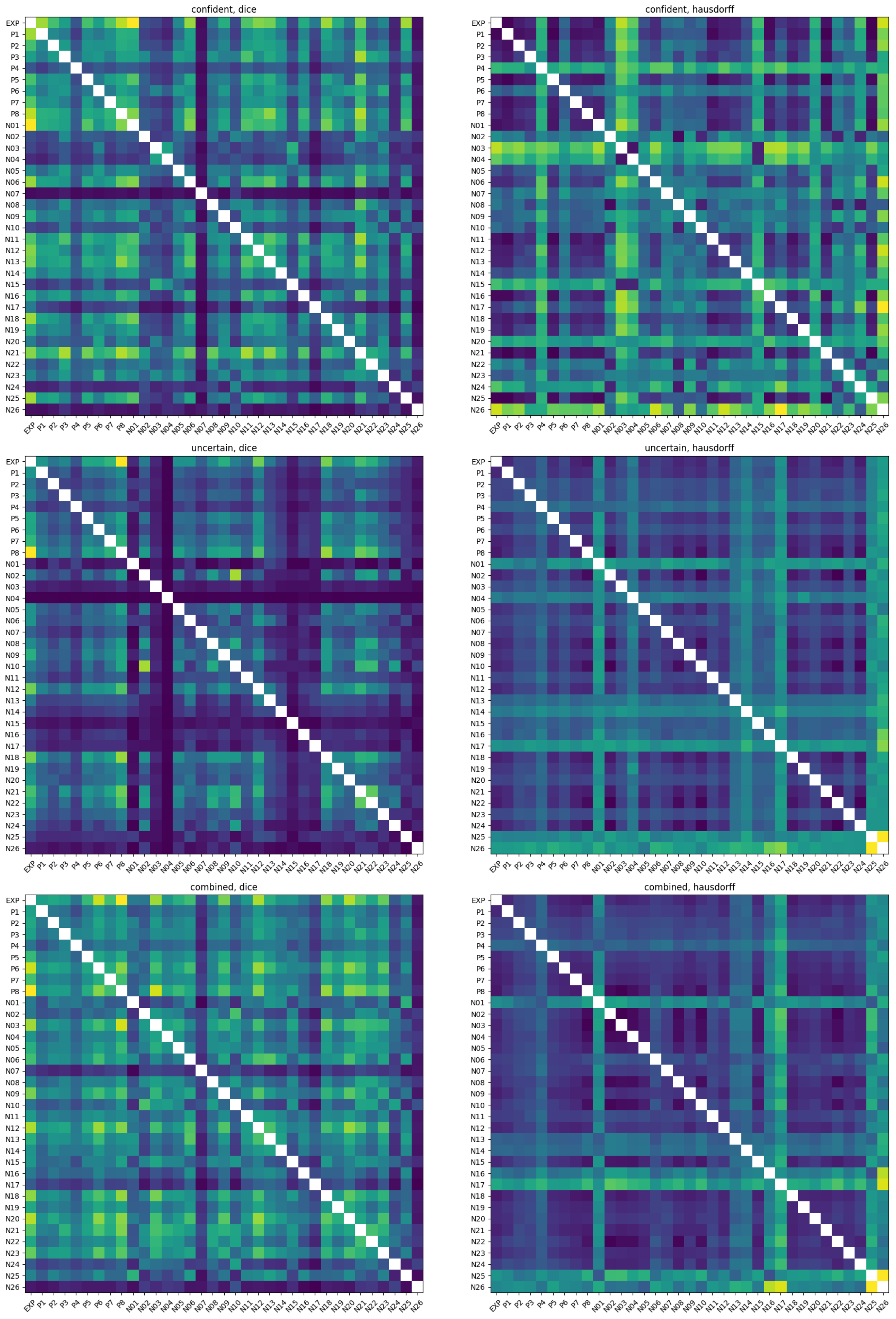}
    \caption{Inter-annotator distances, as measured by Dice coefficient (left) and Hausdorff distance (right). Top row considers only confident fault labels, middle row considers only uncertain fault labels, and bottom row combines both into a single binary label.}
    \label{fig:stats2}
\end{figure}
\section{Annotation Evaluation Metrics and Statistics}
\label{sec:AppD}

\subsection{Metrics}
\label{subsec:Metrics}

\paragraph{Segmentation Task}
For the segmentation task, the main metric used for evaluating the quality of the expert-level fault segmentation predictions is the mean Dice coefficient. This coefficient can be used to compare the pixel-level agreement between two binary masks, in this case the predicted segmentation mask and the ground truth (expert label). The coefficient is given by the following calculation:
\begin{equation}
    D=\frac{2*|X\cap Y|}{|X|+|Y|}
\end{equation}
where $X$ corresponds to the predicted segmentation mask generated by an annotator or the model and $Y$ corresponds to the expert ground truth. 

\paragraph{Detection Task}
For the detection and instance segmentation task, we use the mean average precision (mAP) metric to evaluate the quality of the predictions. mAP is built upon the intersection over union (IoU) metric. The IoU of a prediction mask $X$ and a ground truth mask $Y$ is calculated as
\begin{equation}
    IoU=\frac{X\cap Y}{X \cup Y}
\end{equation}
Calculating the mAP metric involves sweeping over a specified IoU threshold, and calculating an average precision value. At mIOU threshold of 0.5, a predicted mask is counted as correct if its IoU with a ground truth mask that is greater than 0.5.

For each of the possible threshold value $t$, a precision value is calculated using the number of true positives (TP), false negatives (FN), and false positives (FP) that are obtained from comparing the objects in the segmentation and prediction masks:
\begin{equation}
    p=\frac{TP(t)}{TP(t)+FP(t)+FN(t)}
\end{equation}
Finally, the mAP score is calculated as the mean of the above precision values across all IoU thresholds.

\subsection{Statistics}

We gather a number of different statistics to showcase the rich heterogeneity of the \texttt{CRACKS} dataset. Figure \ref{fig:stats1} (top) shows the average number of labels used per image for each annotator, the labeling time for each annotator, and average certainty (taken by calculating the percentage of certain fault pixels with respect to total fault pixels). Figure \ref{fig:stats1} (bottom) showcases the self-consistency values for each individual annotator across the QA images.

We also report the inter-annotator `agreement' or distance in Figure \ref{fig:stats2}. This is calculated by averaging the pairwise distance between each set of labels corresponding to the same image for each possible pair of annotations, using both the Dice coefficient and Hausdorff distance as metrics. We perform this comparison for the certain, uncertain and combined (certain and uncertain) labels, corresponding to the top, middle and bottom rows of the figure respectively. An interesting pattern that seems to emerge in the Hausdorff metric is that when considering only the confident label, the practitioners are far more close to each other than the novices.

\label{subsec:Statistics}
\section{Additional Results}
\label{sec:AppE}
In this section, we provide additional experiments, supplementing the results in Section~\ref{sec:Experiments}. 

\begin{figure}[t]
\centering
\includegraphics[scale =.3]{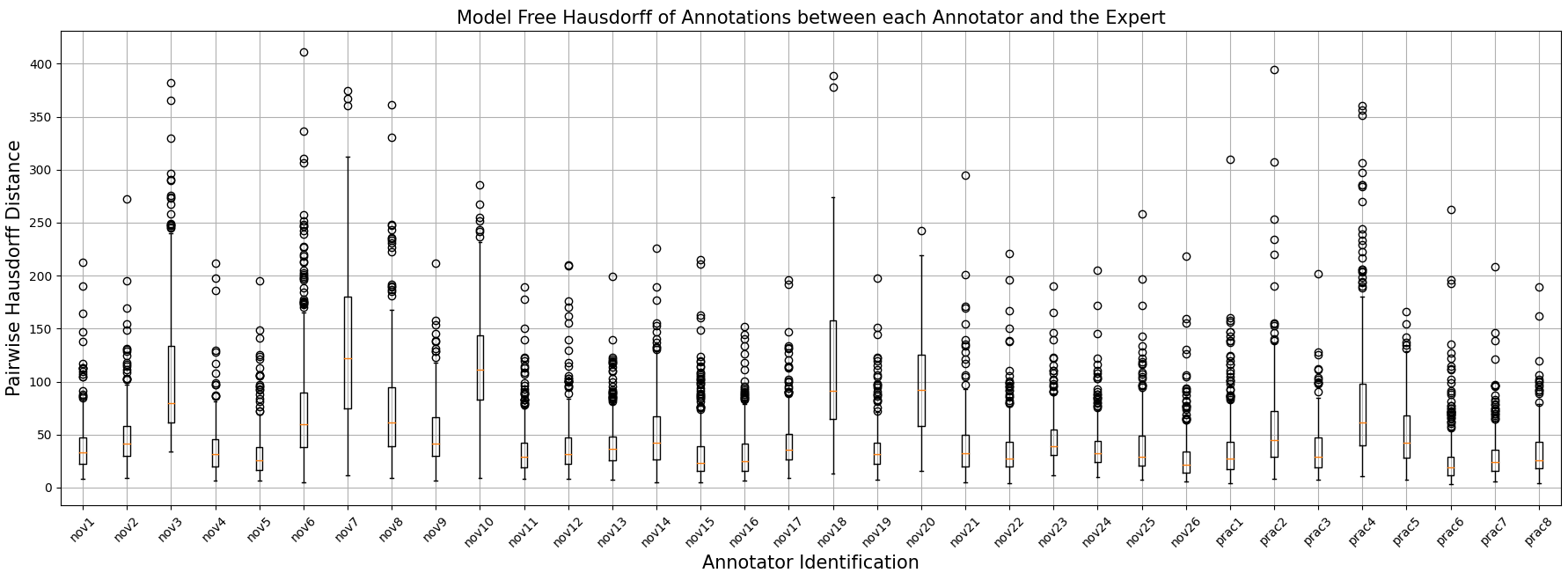}
\vspace{-1.5mm}
\caption{Pairwise Hausdorff between annotations from every practitioner and novice (x-axis) is calculated
against the expert annotation and is presented on the y-axis.}\label{fig:model_free_hausdorff}\vspace{-3mm}
\end{figure}

\begin{figure}[t]
\centering
\includegraphics[scale =.3]{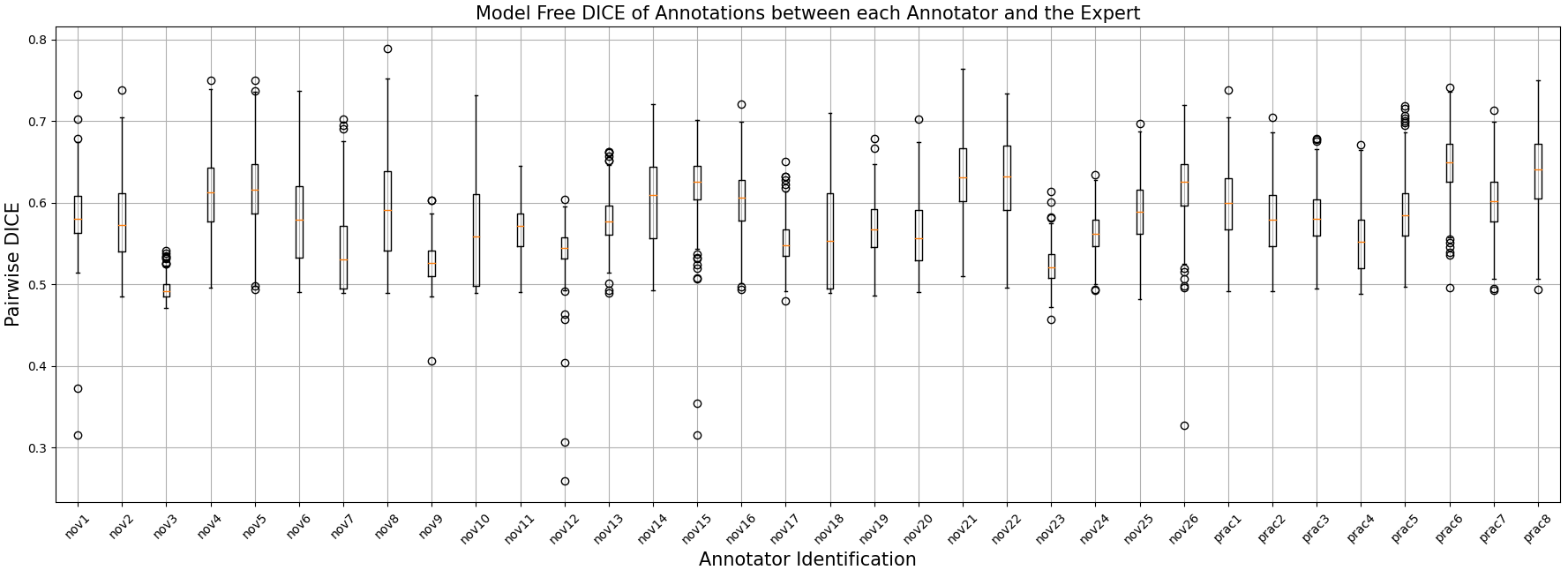}
\vspace{-1.5mm}
\caption{Pairwise DICE between annotations from every practitioner and novice (x-axis) is calculated
against the expert annotation and is presented on the y-axis.}\label{fig:model_free_dice}\vspace{-3mm}
\end{figure}

\begin{figure}[t]
\centering
\includegraphics[scale =.5]{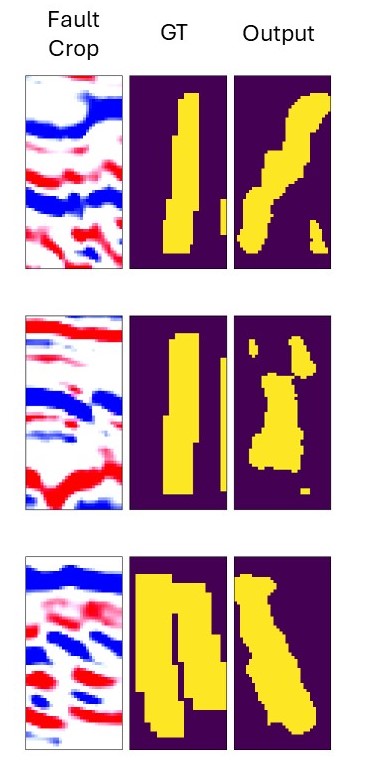}
\vspace{-1.5mm}
\caption{This is a visualization of the fault crops used in the SSL experiments, their associated ground truth, and the output prediction after fine-tuning on one of he annotator label sets.}\label{fig: fault_crop}\vspace{-3mm}
\end{figure}

\subsection{Additional Model Free Metrics}

In Figures~\ref{fig:model_free_hausdorff} and~ \ref{fig:model_free_dice} we show a similar model free comparison between each expert and each annotator (as in Figure \ref{fig:model_free}), with respect to the Hausdorff and DICE metrics respectively. We see that even when using these metrics, the non-expert annotators are able to produce annotations that do not deviate significantly from the expert annotator. This again suggests that these non-expert annotations can potentially be aggregated in an intelligent fashion to approximate the expert's annotations. 

\subsection{SSL Experiments} For the SSL experiments, we use a setup where each fault region is extracted as an object crop and then training and fine-tuning on these object crops rather than the seismic section as a whole. We visualize this version of the dataset in Figure \ref{fig: fault_crop} where we see the image object crop, the associated ground truth, and the predicted output from the model fine-tuned on one of the annotator label sets. We note that the predicted segmentation mask is reasonably close to the ground truth even though the backbone encoder has not been explicitly trained to recognize the presence of faults. This highlights the potential exhibited by SSL algorithms for fault delineation. 

\begin{figure}[t]
\centering
\includegraphics[scale =.3]{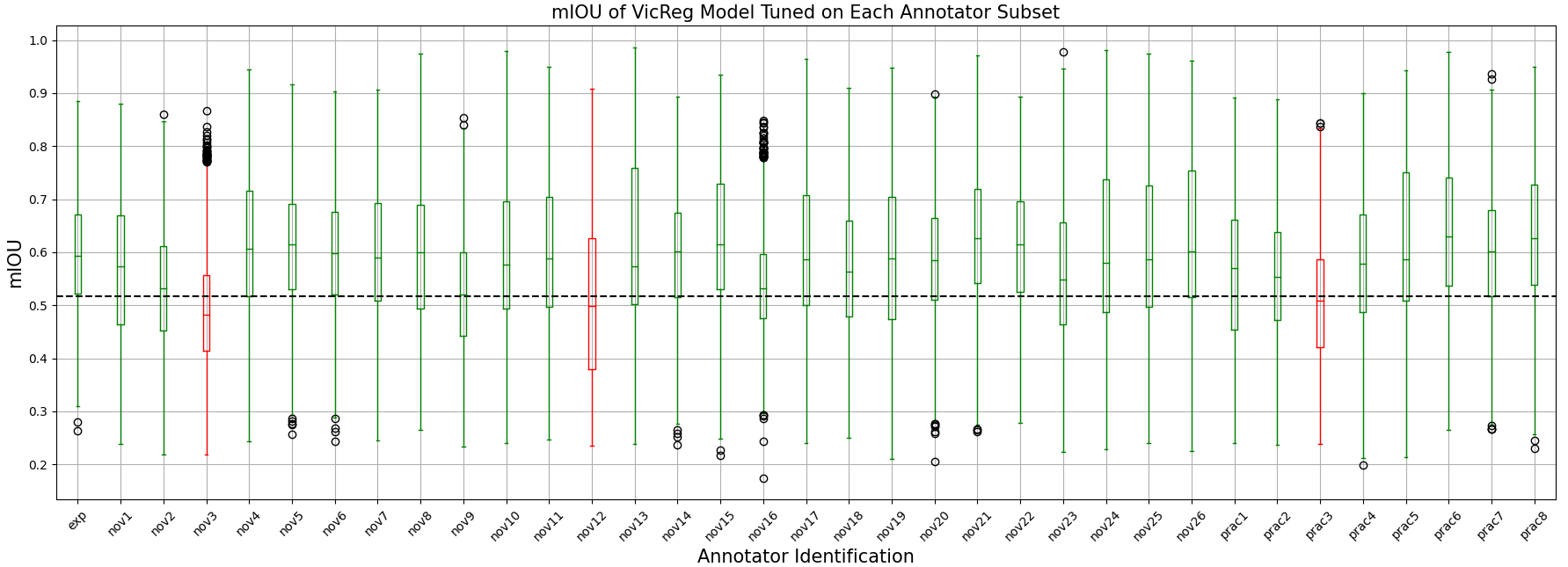}
\vspace{-1.5mm}
\caption{SSL finetuning on each annotator subset using the VicReg methodology.}\label{fig:ssl_vicreg}\vspace{-3mm}
\end{figure}

Additionally, the \texttt{CRACKS} dataset can serve as a useful benchmark for SSL algorithms across a variety of different methods. Therefore, we repeat the same training and testing procedure as before with the VicReg algorithm and show the results in Figure \ref{fig:ssl_vicreg}. We train a ResNet-50 architecture, a LARS optimizer, 100 training epochs, MLP projection head with 1024 nodes across 3 layers, batch size of 64, and the default loss weights found in the original paper. The VicReg algorithm shows a similar trend to the results we visualized in Fig.~\ref{fig:ssl} using the SimCLR algorithm. This shows that SSL can be applied regardless of the type of algorithm that is used.

\subsection{Synthetic Data Comparison} 
\begin{figure}[t]
\centering
\includegraphics[scale =.5]{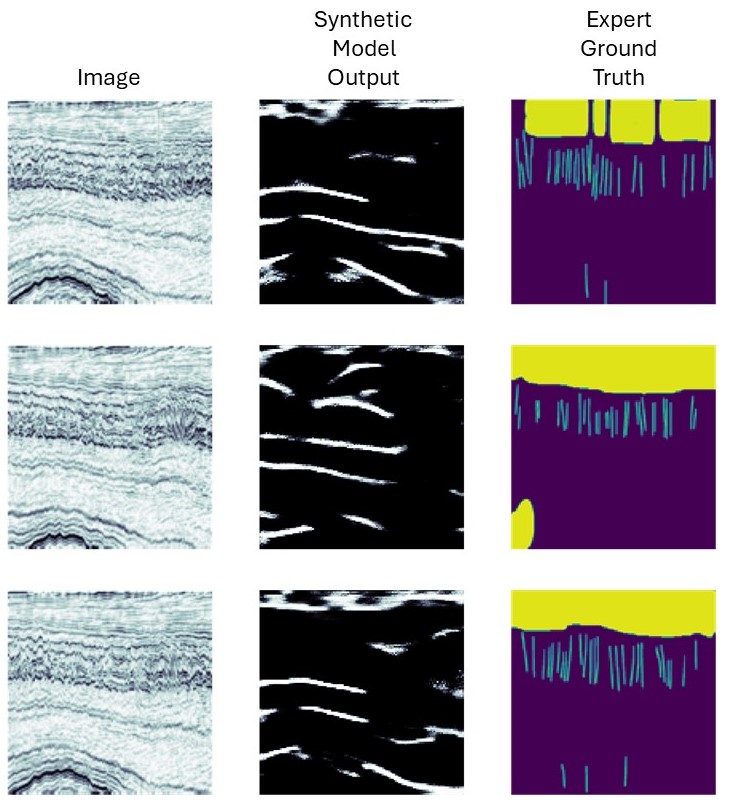}
\vspace{-1.5mm}
\caption{Comparison of ground truth with prediction of model trained on synthetically generated Fault Data.}\label{fig:synthetic}\vspace{-3mm}
\end{figure}

One potential argument against the construction of this dataset is that understanding the interaction of human annotators is unnecessary if we can simply generate high quality labels and examples in a synthetic manner. In this way, the synthetic data will have essentially perfect annotations and it would circumvent the need to study uncertain and noisy human annotations. Recent work \cite{wu2019faultseg3d} showed that synthetically generated data can be used to train and test fault segmentation algorithms. We use their models trained on synthetic data and input the volumes from our dataset. We show some output predictions and compare them with the expert annotated ground truth in Figure \ref{fig:synthetic}. We see that the model trained with synthetic data has no capacity to capture the fine-grained granularity of the faults within our dataset. In fact, it overfits to the presence of different textures within the image, rather than localizing the exact presence of faults through the section itself. Note that seismic data is computed using geological forward models, given wave reflections. Usage of different forward models can alter the seismic data, thereby creating large domain differences. The wide variety of fine-grained changes in fault types make it difficult to formulate a synthetic dataset that can reliably serve as useful annotations for the presence of faults. Additionally, synthetic datasets do not enable a study into the differences in different levels of expertise with respect to the labeling task.

\subsection{Semantic Segmentation Results}
\begin{figure}[t]
    \centering
    \begin{subfigure}[b]{0.3\textwidth}
        \centering
        \includegraphics[width=\textwidth]{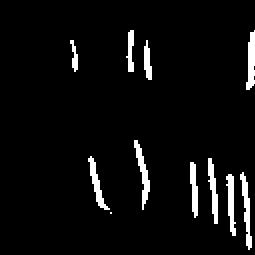}
        \caption{predicted mask using segformer before finetuning}
    \end{subfigure}
    \hfill
    \begin{subfigure}[b]{0.3\textwidth}
        \centering
        \includegraphics[width=\textwidth]{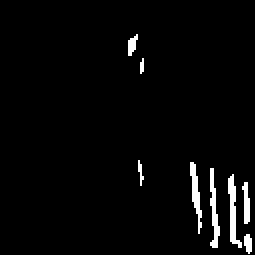}
        \caption{predicted mask using segformer after finetuning}
    \end{subfigure}
    \hfill
    \begin{subfigure}[b]{0.3\textwidth}
        \centering
        \includegraphics[width=\textwidth]{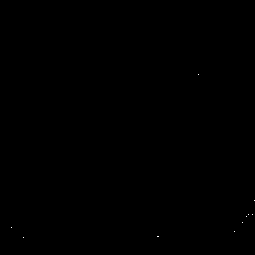}
        \caption{predicted mask using UNet before finetuning}
    \end{subfigure}
    
    \vskip\baselineskip
    
    \begin{subfigure}[b]{0.3\textwidth}
        \centering
        \includegraphics[width=\textwidth]{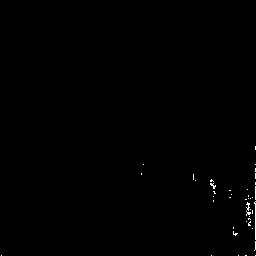}
        \caption{predicted mask using UNet after finetuning}
    \end{subfigure}
    \hfill
    \begin{subfigure}[b]{0.3\textwidth}
        \centering
        \includegraphics[width=\textwidth]{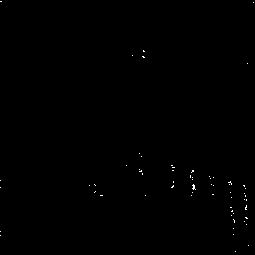}
        \caption{predicted mask using UNet++ before finetuning}
    \end{subfigure}
    \hfill
    \begin{subfigure}[b]{0.3\textwidth}
        \centering
        \includegraphics[width=\textwidth]{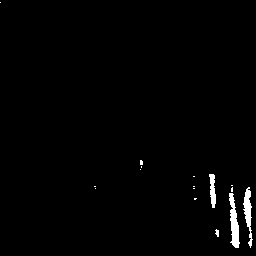}
        \caption{predicted mask using UNet++ after finetuning}
    \end{subfigure}
    
    \vskip\baselineskip
    
    \begin{subfigure}[b]{0.3\textwidth}
        \centering
        \includegraphics[width=\textwidth]{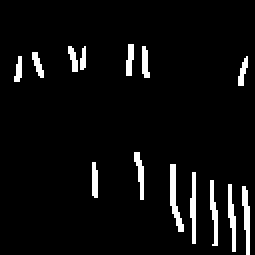}
        \caption{The ground truth labeled by the expert}
    \end{subfigure}

    \caption{The predicted masks using segformer, Unet, and Unet++ before and after the expert finetuning, against the ground truth labeled by the expert.}
    \label{fig:segresults}
\end{figure}

In this section, we augment Table~\ref{table:segmentation_results_beforefine} by training SOTA models using three different seeds. The SOTA models include segformer, Deeplab v3 (ResNet 50), Unet, and Unet++. We use the same setup and training settings as our previous segmentation experiment. The results are shown in \ref{table:segmentation_results_sub}. We qualitatively provide the predicted masks for each model against the ground truth labeled by the expert in Figure \ref{fig:segresults}. Note that section-wise segmentation is far harder than object crops in SSL from Fig.~\ref{fig: fault_crop}.
\begin{table}[h!]
\begin{center}
\begin{scriptsize}
\begin{sc}
\caption{Averaged novice and practitioner Section-wise Fault Segmentation Results before and after finetuning on expert. Hausdorff distance shows significant decrease (in red) while all other changes are statistically insignificant. These are the results of training the models multiple times using different seeds.}
\begin{tabular}{lcccccr}
\toprule
Models & \multicolumn{3}{c}{Before Finetuning} & \multicolumn{3}{c}{After finetuning}\\ 
& mIOU & Hausdorff & DICE Score & mIOU & Hausdorff & DICE Score \\
\midrule
Segformer \cite{xie2021segformer}                          & 0.53 ± 0.003  & 42.4 ± 0.6 & 0.58 ± 0.0005 & 0.54 ± 0.002 & \color{red}{59.57 ± 0.52} & 0.56 ± 0.004  \\
Unet \cite{ronneberger2015u}                            & 0.49 ± 0.006  & 69.09 ± 4.68  & 0.5 ± 0.02 & 0.49 ± 0.006 & \color{red}{90.36 ± 1.66} & 0.5 ± 0.007  \\
Unet++ \cite{zhou2019unetplusplus}                     & 0.49 ± 0.004  & 69.44 ± 5.27  & 0.51 ± 0.009 & 0.49 ± 0.0012  & \color{red}{92.8 ± 5.28} & 0.51 ± 0.001  \\
Deeplab v3 \cite{chen2017deeplab}   & 0.48 ± 0.002  & 56.8 ± 2.13  & 0.52 ± 0.002 & 0.48 ± 0.0002 & \color{red}{123.9 ± 1.57} & 0.517 ± 0.0003  \\
\bottomrule
\end{tabular}
\vspace*{.1mm}
\label{table:segmentation_results_sub}
\end{sc}
\end{scriptsize}
\end{center}
\end{table}

\begin{figure}[t]
\centering
\includegraphics[scale =.6]{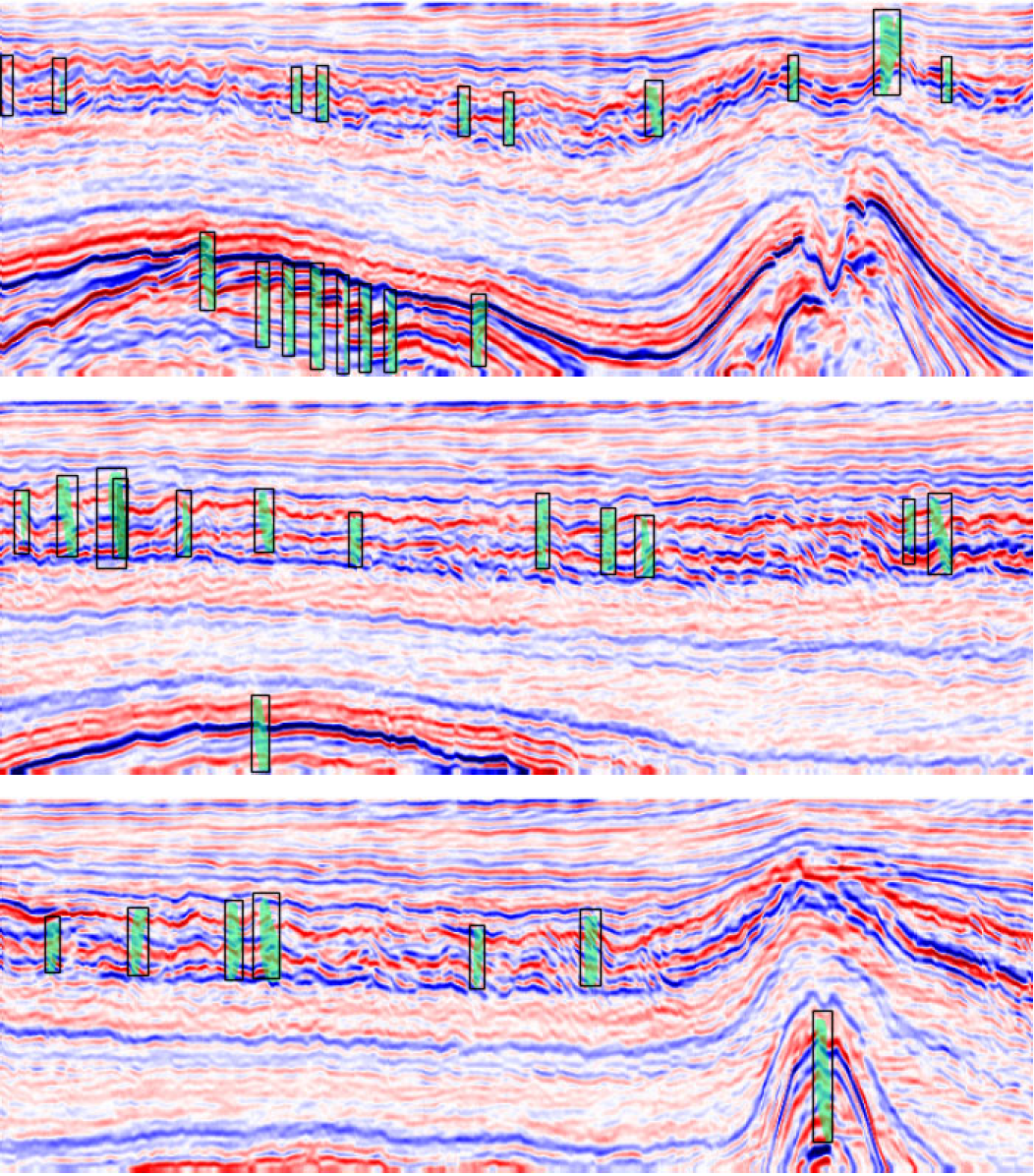}
\vspace{-1.5mm}
\caption{Prediction results of 3 sample sections for YOLOv5 instance segmentation}\label{fig:predictionsamplesyolov5segmentation}\vspace{-3mm}
\end{figure}

\subsection{Label Fusion based on confidence and expertise}

In Figure \ref{fig:predictionsamplesyolov5segmentation}, we show sample prediction results for 3 different sections for YOLOv5 instance segmentation. Next, we fuse the confidence and expertise labels in the following fashion to obtain results:

\begin{itemize}
    \item In Table \ref{table:labelfusioncertain}, we show the results for label fusion between uncertain and certain labels. For this fusion, we use labels from both practitioners and novices, with certain labels assigned a weight 1.5 times greater than that of uncertain labels. 
    \item We augment rows 6 and 7 in Table~\ref{yolodetectionresults} by utilizing an alpha parameter to fuse novice and practitioner labels. The alpha parameter controls the weighting between the contributions of novices and practitioners, calculated as: $(\alpha \times Novice + (1-\alpha) \times Practitioner) $. In Table \ref{table:labelfusionalpha}, we show the results for label fusion between practitioners and novices as a function of various alpha parameters.  
\end{itemize}

\begin{table}[t]
\begin{center}
\begin{tiny}
\begin{sc}
\caption{Results of fault detection and instance segmentation for label fusion between certain and uncertain labels using mean Average Precision at 0.5. }
\begin{tabular}{lcr}
\toprule
Training Set  & \multicolumn{1}{c}{Detection} & Instance Segmentation  \\
\midrule

Certain-Uncertain Label Fusion     & 0.308 ± 0.004   &      0.458 ± 0.002      \\

Certain-Uncertain Label Fusion (expert fine-tuned)    & 0.425 ± 0.004   &      0.168 ± 0.002      \\
\bottomrule
\end{tabular}
\vspace*{.1mm}
\label{table:labelfusioncertain}
\end{sc}
\end{tiny}
\end{center}
\end{table}

\begin{table}[t]
\begin{center}
\begin{tiny}
\begin{sc}
\caption{Results of fault detection and instance segmentation for label fusion between practitioner and novices with different alpha parameters $(\alpha \times Novice + (1-\alpha) \times Practitioner) $ using mean Average Precision at 0.5.  }
\begin{tabular}{lcr}
\toprule
Training Set Label Fusion Alpha Parameter   & \multicolumn{1}{c}{Detection} & Instance Segmentation  \\
\midrule

Label Fusion $\alpha=0$     & 0.342  &    0.118     \\
Label Fusion $\alpha=0.2$     & 0.329  &    0.0535     \\
Label Fusion $\alpha=0.4$     & 0.325  &   0.0433      \\
Label Fusion $\alpha=0.5$     & 0.328  &     0.0813    \\
Label Fusion $\alpha=0.6$     & 0.336  &    0.0892     \\
Label Fusion $\alpha=0.8$     & 0.324  &   0.00249      \\
Label Fusion $\alpha=1$     & 0.317  &    0.0808     \\
\midrule
Label Fusion $\alpha=0$  (expert fine-tuned)     &  0.457 &  0.16     \\
Label Fusion $\alpha=0.2$  (expert fine-tuned)    & 0.465  &   0.154  \\
Label Fusion $\alpha=0.4$  (expert fine-tuned)    & 0.458  &    0.145     \\
Label Fusion $\alpha=0.5$  (expert fine-tuned)    & 0.465  &  0.168       \\
Label Fusion $\alpha=0.6$  (expert fine-tuned)     & 0.445  &      0.149   \\
Label Fusion $\alpha=0.8$  (expert fine-tuned)    & 0.449  &     0.134    \\
Label Fusion $\alpha=1$  (expert fine-tuned)    & 0.438  &     0.143    \\

\bottomrule
\end{tabular}
\vspace*{.1mm}
\label{table:labelfusionalpha}
\end{sc}
\end{tiny}
\end{center}
\end{table}

\subsection{Computational Resources}
\label{app: cluster}
All experiments were run on PCs with two NVIDIA GeForce GTX TITAN X 12 GB GPUs.

\end{document}